\documentclass[conference]{IEEEtran}
\usepackage{times}

\usepackage[numbers]{natbib}
\usepackage{multicol}
\usepackage[bookmarks=true]{hyperref}
\usepackage{graphicx}

\usepackage{multirow}
\usepackage{booktabs}
\usepackage{amssymb} 
\usepackage{color}
\usepackage{makecell}
\usepackage{caption} 
\usepackage{graphicx}
\usepackage{xcolor}
\usepackage{amsmath}
\newcommand{\ours}{HumDex}

\usepackage{caption} 

\begin{document}

\title{\ours:\\Humanoid Dexterous Manipulation Made Easy}

\author{%
  Liang Heng$^{*1, 2}$ \quad Yihe Tang$^{* 1}$ \quad Jiajun Xu$^{2}$ \quad Henghui Bao$^{2}$ \quad Di Huang$^{2}$ \quad Yue Wang$^{1}$ \\
  $^{1}$USC Physical Superintelligence (PSI) Lab \quad $^2$WorldEngine AI\\
  $^*$Equal Contribution}


\twocolumn[{%
  \renewcommand\twocolumn[1][]{#1}%
  \maketitle 
  \begin{center}
    \centering
    \includegraphics[width=0.98\textwidth]{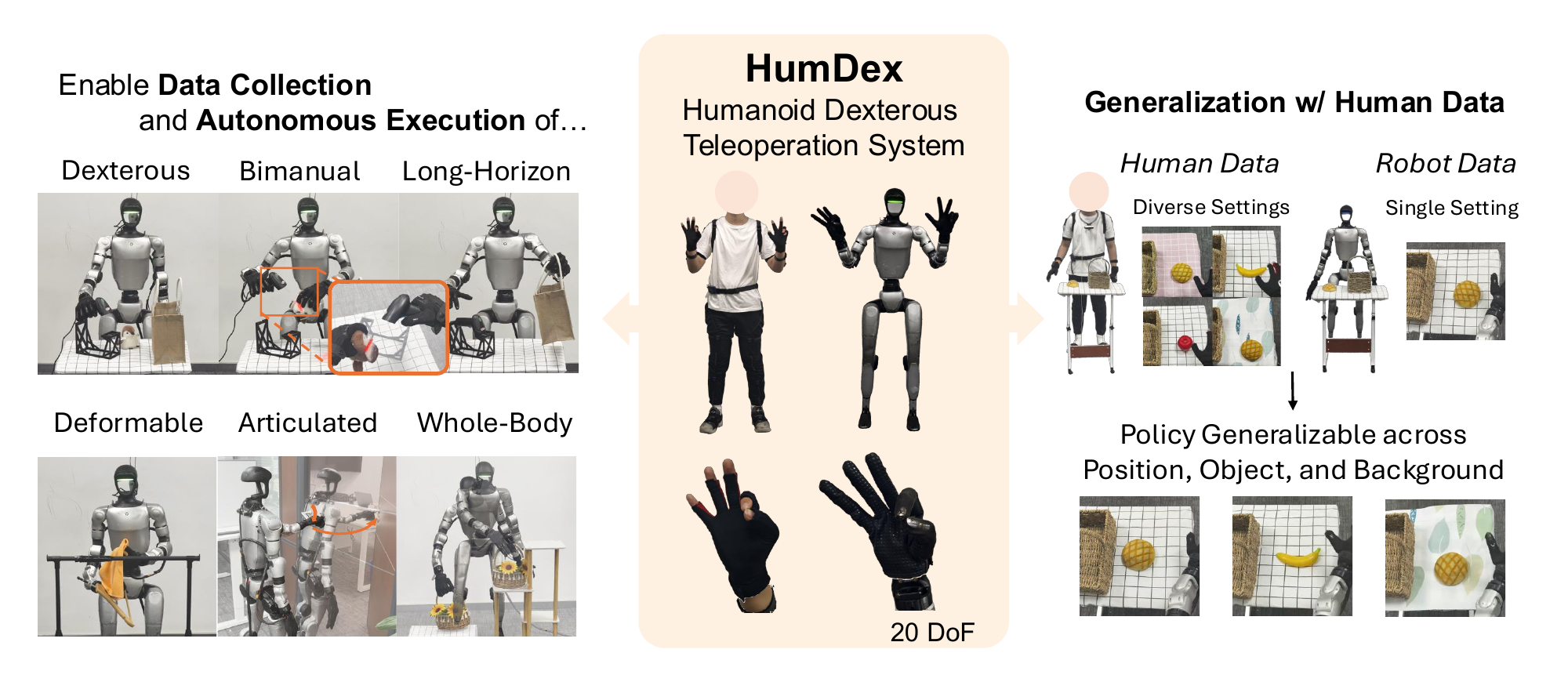} 
    \vspace{-0.1in} 
    
    \captionof{figure}{\textbf{The HumDex System.} Our portable teleoperation system enables efficient collection of high-quality dexterous manipulation data. Left: We demonstrate data collection and autonomous policy execution on challenging tasks featuring dexterous manipulation, bimanual coordination, long-horizon planning, deformable and articulated object manipulation, and whole-body movement. Middle: We use a Unitree-G1 humanoid and two 20 DoF dexterous hands. Right: By pretraining robot policy on diverse human data, our policy generalizes to new positions, objects, and backgrounds unseen in robot data.}
    \label{fig:teaser}
  \end{center}%
  \vspace{1em} 
}]

\begin{abstract}
This paper investigates humanoid whole-body dexterous manipulation, where the efficient collection of high-quality demonstration data remains a central bottleneck.
 Existing teleoperation systems often suffer from limited portability, occlusion, or insufficient precision, which hinders their applicability to complex whole-body tasks.  To address these challenges, we introduce \ours, a portable teleoperation system designed for humanoid whole-body dexterous manipulation.
Our system leverages IMU-based motion tracking to address the portability-precision trade-off, enabling accurate full-body tracking while remaining easy to deploy. For dexterous hand control, we further introduce a learning-based retargeting method that generates smooth and natural hand motions without manual parameter tuning. Beyond teleoperation, \ours~  enables efficient collection of human motion data. Building on this capability, we propose a two-stage imitation learning framework that first pre-trains on diverse human motion data to learn generalizable priors, and then fine-tunes on robot data to bridge the embodiment gap for precise execution. We demonstrate that this approach significantly improves generalization to new configurations, objects, and backgrounds with minimal data acquisition costs. The entire system is fully reproducible and open-sourced at \url{https://github.com/physical-superintelligence-lab/HumDex}. 
\end{abstract}

\IEEEpeerreviewmaketitle

\section{Introduction}
Humanoid dexterous manipulation holds great promise for unlocking robots to perform complex, long-horizon loco-manipulation tasks in the real world. Current robotic systems often resort to imitation learning~\cite{heng2025imagine2act, zhang2025molevladynamiclayerskippingvision, li20253ds, Li_2025_CVPR, li2025selfcorrectingvisionlanguageactionmodelfast} that has shown great success in acquiring complex manipulation skills.  
These methods rely heavily on high-quality task demonstration data collected through costly robot teleoperation. However, acquiring such data for humanoid robots with dexterous hands remains a critical bottleneck due to their complex morphology.

While huge progress has been made on table-top robot data collection~\cite{zhao2023learningfinegrainedbimanualmanipulation, 10801581, heng2025rwor}, teleoperation systems for humanoid robots and dexterous hands are way less mature. Previous efforts with different hardware solutions exhibit their own limitation and trade-off. Motion-capture-based~\cite{ze2025twist} (e.g., optical tracking) or exoskeleton-based~\cite{ben2025homie} systems can achieve high accuracy but require fixed infrastructure, which severely limits the environments in which data can be collected. In contrast, VR-based alternatives~\cite{ze2025twist2, li2025amo, li2025clone, luo2025sonic} offer greater portability but suffer from reduced accuracy and occlusion issues. For instance, operators’ hands must remain within the sensors’ field of view to maintain tracking stability, constraining the range of feasible motions and, consequently, the set of tasks that can be demonstrated. Furthermore, despite recent advances in humanoid motion retargeting and low-level locomotion policies~\cite{ze2025twist, ben2025homie, ze2025twist2, li2025amo, li2025clone, luo2025sonic}, dexterous hand control still largely relies on optimization-based retargeting, leading to reduced accuracy and limited generalization.

In this work, we introduce \ours (Fig.~\ref{fig:teaser}), a portable motion-tracker-based teleoperation system for whole-body dexterous manipulation. Our system addresses the portability-precision trade-off by leveraging IMU-based tracking, enabling high-precision tracking while maintaining portability. For dexterous hand control, we propose a learning-based retargeting system trained on collected teleoperation data, which produces smooth and natural hand motions without manual parameter tuning. This hand retargeting method, compared to previous optimization-based alternatives, achieves significantly better performance in real-world deployment. We demonstrate the effectiveness of \ours\; on a suite of challenging tasks involving whole-body motion, bimanual coordination, and fine-grained dexterous manipulation. Overall, our system enables faster demonstration collection, higher success rates, and improved data quality relative to existing approaches.

Beyond teleoperation, our tracking system also enables efficient collection of human data of the same tasks, which offers better collection efficiency than teleoperation, thus serves as an additional data source for pre-training or co-training. However, due to embodiment gaps, directly retargeting human motion to the humanoid leads to inaccurate movements, which often leads to manipulation failures. Consequently, prior works on tabletop dexterous manipulation performs alignment~\cite{kareer2025egomimic, qiu2025humanoid} or correction strategies~\cite{wang2024dexcap} to mitigate this gap, while those on humanoid manipulation rely solely on teleoperation data~\cite{li2025amo, ze2025twist2, heng2025vitacformerlearningcrossmodalrepresentation}. To effectively leverage the diversity and motion prior in human data without explicit alignment,  we propose a two-stage imitation learning framework. First, we train the policy on human demonstrations collected in diverse settings, where we use retargeting results as a joint target, and approximate proprioceptive states with previous action. Then, we fine-tune on robot teleoperation data only, refining movements towards the robot embodiment. As shown in Table \ref{tab:human_generalize}, our approach achieves successful task execution while retaining generalization to new object positions, categories, and backgrounds without requiring robot data under those settings.

In conclusion, our contributions are:  (1) a portable and efficient teleoperation system for humanoid dexterous manipulation, (2) a learning-based hand retargeting method, and (3) a two-stage training pipeline that leverages human data to improve generalization while reducing the need for teleoperation data.

\begin{figure*}[t]
  \centering
  \includegraphics[width=\textwidth]{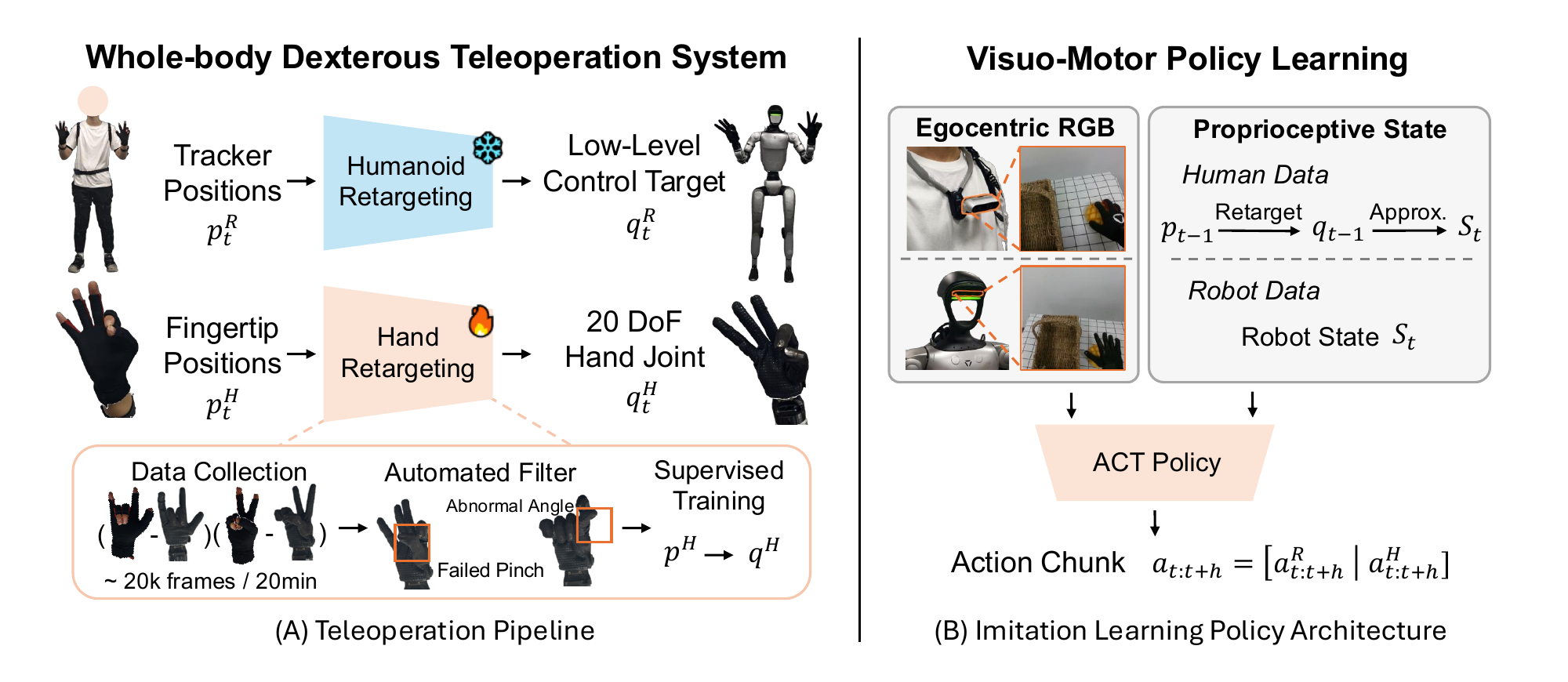} 
  \caption{\textbf{System Overview.} (A) Our teleoperation pipeline and hand retargeting policy training. (B) Our imitation learning policy architecture. We approximate proprioceptive states missing in human data with previous-frame actions.}
  \label{fig:method_overview}
\end{figure*}

\section{Related Works}

\subsection{Humanoid Whole-Body Dexterous Teleoperation}
Existing humanoid whole-body teleoperation systems can be categorized by their tracking hardware. Motion-capture-based~\cite{ze2025twist} and exoskeleton-based~\cite{ben2025homie} systems achieve high tracking accuracy but suffer from portability issues — mocap requires a dedicated room setup, while exoskeletons are heavy and typically require seated operation. Vision-based and VR-based systems~\cite{ze2025twist2, li2025amo, luo2025sonic, li2025clone} offer better portability but suffer from occlusion: operators must keep their hands visible at all times, restricting feasible motions. In this work, we adopt IMU-based motion tracking, which consists of only 15 lightweight trackers worn on the body, providing unconstrained motion capture with high tracking quality.
Beyond hardware, we also investigate a more challenging robot configuration. Prior whole-body teleoperation works employ simplified end-effectors such as parallel grippers or three-fingered hands, often controlled via binary open/close signals (e.g., VR controller triggers). Consequently, demonstrated tasks are limited to simple object interactions such as pick-and-place. In contrast, our system supports full dexterous control of a 20-DoF hand, enabling fine-grained manipulation such as grasping a handheld barcode scanner and pulling its trigger, while simultaneously supporting whole-body movement.

\subsection{Dexterous Hand Retargeting}
Dexterous hand retargeting is a key component of teleoperation driven demonstration collection system. It maps human hand features to a robot hand under substantial embodiment gaps. A common approach is optimization-based policy, which formulates the mapping as a constrained inverse kinematics (IK) or nonlinear least squares problem. In these formulations, objective terms typically preserve task-relevant geometric relations, while constraints enforce robot executability. To improve stability in teleoperation and contact-rich manipulation, many methods additionally incorporate temporal-consistency regularizers and contact-consistency/interpenetration penalties to discourage implausible hand and object penetrations and stabilize interaction~\cite{dexmachinafunctional,dexflow, keyobjectives}.

In contrast, learning-based approaches predict robot hand configurations directly from human observations, reducing reliance on hand-crafted objectives and enabling constant inference. GeoRT~\cite{geort} proposes an ultrafast neural retargeting approach guided by principled geometric criteria, achieving real-time performance without test-time optimization and supporting scalable teleoperation pipelines~\cite{geort,dexteritygen}.

Our approach follows this learning-based direction with a lightweight supervised formulation. Given the 3D positions of five fingertips, we train a small MLP regressor to predict robot hand joint angles on paired fingertip--joint samples. 

\subsection{Imitation Policy Learning with Human Data}
Imitation learning has advanced rapidly, with several works leveraging human data for robot learning. Some approaches use egocentric human video as a diverse source for visual representation learning~\cite{nair2022r3m,xiao2022masked, radosavovic2023real}. For dexterous manipulation specifically, prior works focus primarily on fixed-base arms and tabletop tasks, bridging the human-robot gap through hardware design that mimics human embodiment~\cite{kareer2025egomimic}, human-in-the-loop correction~\cite{wang2024dexcap}, or aligning action space~\cite{qiu2025humanoid}.

However, for humanoid robots, the embodiment gap to humans is substantially larger and cannot be easily addressed through hardware alignment. While human motion data has been widely used for learning humanoid low-level control and general whole-body movements, these works emphasize approximate motion following rather than the precision required for manipulation task success. Consequently, manipulation tasks with humanoids still rely on teleoperated robot data~\cite{ben2025homie, ze2025twist2, li2025amo}.

Our approach investigates leveraging human data for humanoid loco-manipulation with dexterous hands. Through our two-stage training framework---pretraining on human demonstrations followed by finetuning on robot data---we achieve improved generalization without requiring special embodiment alignment or extensive data post-processing.


\section{Method}

In this section, we present \textbf{\ours} (Fig.~\ref{fig:method_overview}), a portable teleoperation system for humanoid whole-body dexterous manipulation, along with a two-stage learning framework to leverage human motion data. The system overview is illustrated in Fig.~\ref{fig:method_overview}. Sec~\ref{sec:method_teleop} describes our IMU-based teleoperation system for efficient whole-body data collection. Section~\ref{sec:method_hand} presents our learning-based approach for dexterous hand retargeting. Section~\ref{sec:method_learning} outlines our imitation learning pipeline and two-stage training strategy that pretrains on human data before finetuning on robot demonstrations.

\subsection{Whole-Body Dexterous Teleoperation System} \label{sec:method_teleop}

\textbf{Problem Formulation} 
We address the problem of real-time, whole-body dexterous teleoperation within a unified control framework. Our goal is to map the motion of a human operator, captured via a wearable IMU-based system, to a humanoid robot to execute diverse loco-manipulation tasks. Following the hierarchical architecture proposed in TWIST2~\cite{ze2025twist2}, we decouple the system into a task-agnostic low-level controller $\pi_{low}$ and a high-level command generator $\pi_{high}$.

\subsubsection{Low-Level Hierarchical Control Interface}
To achieve stable whole-body control while enabling precise dexterous manipulation, we adopt a modular control strategy. We leverage the robust general motion tracking policy from TWIST2~\cite{ze2025twist2} to handle the robot's base and body balancing, while integrating our custom retargeting pipeline for the dexterous hands.

\paragraph{Body Controller Interface} 
We utilize the pre-trained motion tracking policy $\pi_{low}$ from TWIST2~\cite{ze2025twist2} as the foundation for body stability. Consistent with the standard interface, the whole-body reference command vector $p_{cmd}$ is defined as:
\begin{equation}
    p_{cmd} = [\mathbf{v}_{root}, z_{ref}, \mathbf{\Theta}_{root}, \dot{\psi}_{ref}, q_{ref}]^\top
\end{equation}
where $q_{ref} \in \mathbb{R}^{N}$ represents the whole-body joint targets. 
Crucially, we conceptually decompose the joint targets into two distinct components for execution: $q_{ref} = [q_{body}, q_{hand}]$. The body component $q_{body}$ is fed into the policy $\pi_{low}$ to generate actuation torques for dynamic balancing and locomotion, while $q_{hand}$ is directly executed via joint position controllers to track precise dexterous motions. 

Beyond TWIST2~\cite{ze2025twist2}, our framework also maintains compatibility with other low-level controllers such as SONIC~\cite{luo2025sonic}. The reference command $q_{body}$ is efficiently streamed to these controllers via ZMQ messaging backbone, ensuring low-latency synchronization between the high-level retargeting and low-level execution.

\paragraph{Dexterous Hand Control} 
Unlike TWIST2, which simplifies the hand control to a binary open-close mechanism, we implement a fine-grained dexterous retargeting module. Specifically, we train a lightweight MLP regressor that maps the 3D positions of the operator's five fingertips (captured via IMU gloves) directly to the robot's 20-DoF hand joint angles. This learning-based approach ensures smooth and natural motion reconstruction without manual parameter tuning. The computed hand targets $q_{hand}$ are then concatenated with the body targets $q_{body}$ to form the unified $q_{ref}$ in Eq.~(1).

\subsubsection{High-Level Teleoperation via IMU-based Retargeting} 
In the teleoperation phase, the high-level policy $\pi_{high}$ is composed of the human operator and a retargeting algorithm. Unlike prior works constrained by VR-based tracking~\cite{ze2025twist2} or bulky optical motion capture, we utilize a custom IMU-based wearable interface designed for minimal intrusiveness. The system is lightweight and unobtrusive, ensuring the operator's motion remains natural and unencumbered. Furthermore, it supports continuous operation exceeding 10 hours and maintains robust connectivity over a range of 50+ meters, significantly expanding the data collection envelope beyond room-scale setups.

To map the IMU-derived skeleton $\mathcal{S}_H$ to robot commands, we employ the General Motion Retargeting (GMR) framework~\cite{araujo2025retargeting, ze2025twist}. Given that IMU-based systems inherently suffer from global position drift, we adopt a \textit{pelvis-centric} optimization formulation. Specifically, we solve for the robot configuration $q^*$ that minimizes the orientation error of all links and the \textit{relative} position error of key end-effectors:

\begin{equation}
\begin{split}
    q^*(t) = \arg\min_{q} \bigg( & \sum_{i \in \mathcal{L}_R} w_i^R \|R_i^H(t) - R_i^R(q)\|^2 \\
    & + \sum_{j \in \mathcal{L}_P} w_j^P \|p_j^{H, \text{rel}}(t) - p_j^{R, \text{rel}}(q)\|^2 \bigg)
\end{split}
\end{equation}

where $\mathcal{L}_R$ and $\mathcal{L}_P$ denote the sets of links for orientation and position tracking, respectively. Crucially, $p_j^{H, \text{rel}}$ represents the position of the $j$-th end-effector relative to the operator's pelvis frame, rather than the absolute world frame. This formulation~\cite{ze2025twist2} allows us to leverage the standard GMR solver to achieve robust whole-body tracking without relying on drift-prone absolute position measurements.

\subsubsection{Hardware Setup} 
To ensure robustness against occlusion and achieve true ``in-the-wild" portability, our system utilizes a fully tether-free inertial motion capture interface. The hardware configuration is divided into whole-body tracking and dexterous hand tracking.

\paragraph{Body Tracking Interface} 
Our framework is compatible with diverse inertial motion capture solutions. We primarily utilize a commercial 15-node system placed on standard kinematic chains, Vdmocap, to ensure high-precision baseline tracking. 
To further demonstrate the accessibility and scalability of our approach, we implemented a highly cost-effective (\textless \$200) custom wearable system based on the open-source SlimeVR ecosystem~\cite{slimevr}. 
We adopted the architecture from the official hardware guide, allowing for flexible configurations ranging from 6 to 16 nodes to dynamically adjust setup complexity.

\noindent\textbf{Sensor and Electronics.} Each custom node is powered by an ICM-45686 6-axis IMU. We specifically selected this high-performance sensor for its superior precision and low gyroscope drift, ensuring the tracking fidelity required for fine-grained teleoperation. To ensure stable connectivity in complex Wi-Fi environments (e.g., crowded labs), we adopted the non-standard wireless communication scheme supported by the SlimeVR software. 
Specifically, we utilize the Nordic nRF52840 SoC running the Enhanced ShockBurst (ESB) protocol, which significantly reduces packet loss and interference compared to standard UDP-over-WiFi solutions.

\noindent\textbf{Form Factor and Efficiency.} 
The trackers are assembled for minimal intrusiveness. Each node weighs less than 20g and is encased in a compact 3D-printed housing. 
By leveraging the optimized power management features of the ecosystem, the system achieves over 20 hours of continuous operation on a single charge, supporting extensive data collection sessions. 
The raw IMU data is fused on a central receiver dongle to reconstruct the full human skeleton $\mathcal{S}_H$.

\paragraph{Dexterous Hand Interface.} 
For fine-grained manipulation, we leverage high-fidelity commercial inertial gloves—supporting both Vdhand and Manus—for hand motion capture. Each glove utilizes an array of IMUs distributed across the dorsum and finger segments. Unlike vision-based tracking, which suffers from severe self-occlusion during object interaction, this pure inertial solution ensures that complex grasping motions are accurately captured even when the hands are obstructed by objects or the robot's body.


\begin{figure*}[t]
  \centering
  \includegraphics[width=\textwidth]{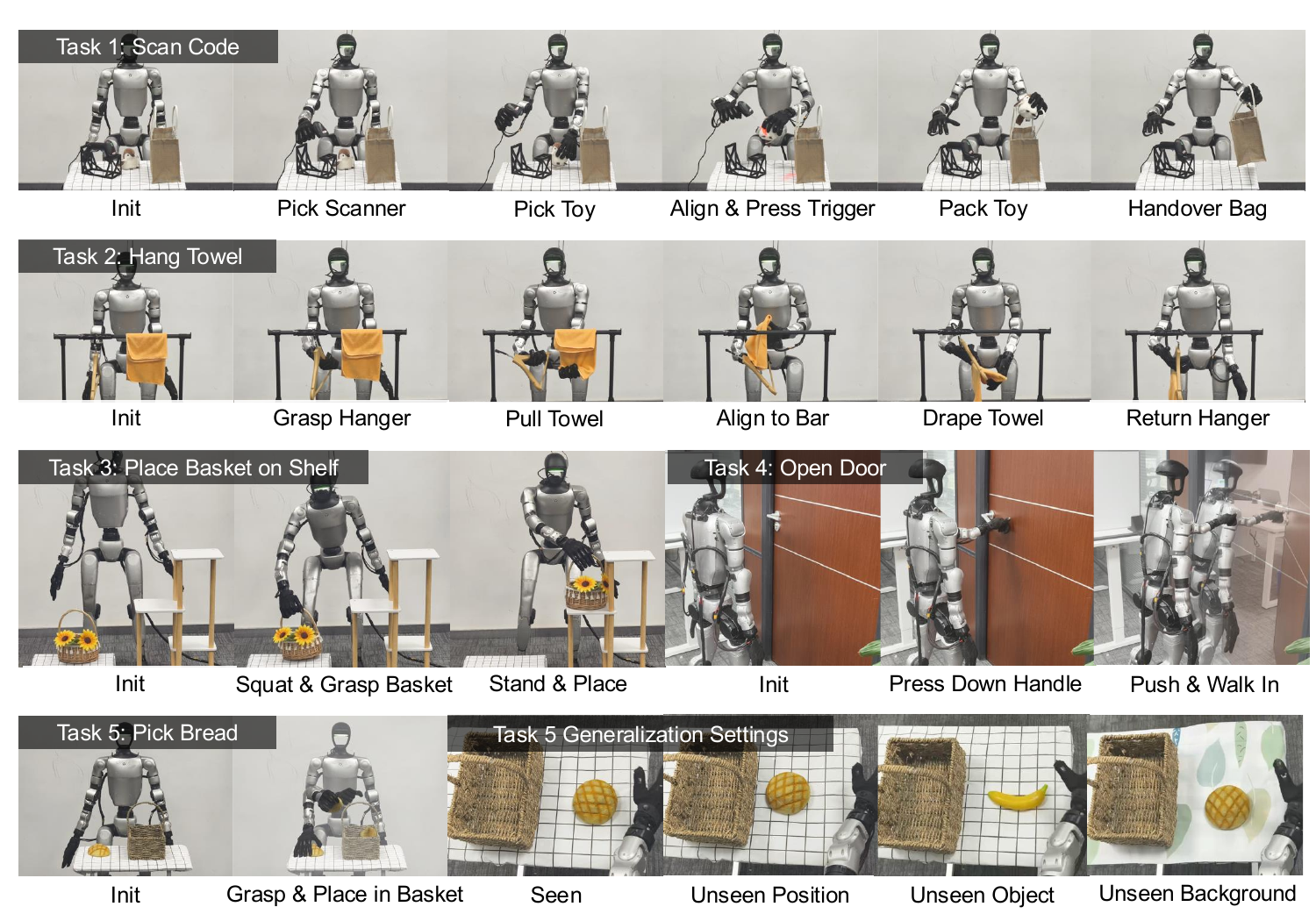} 
  \caption{\textbf{Evaluation Tasks and Generalization.} We visualize the initial state and key steps in our evaluated tasks. In the Task 5 generalization test, robot data used for training only consists of the Seen (position, object, background) setting. }
  \label{fig:tasks}
\end{figure*}

\subsection{Learning-Based Dexterous Hand Retargeting.} 


\label{sec:method_hand}
As mentioned above, our teleoperation setup uses inertial gloves for hand tracking. At each timestep, the glove provides the 3D positions of five fingertips (thumb, index, middle, ring, and little). We express these fingertip positions in the glove wrist frame to avoid sensitivity to global drift. The goal is to map this human fingertip observation to executable robot hand joint targets for a 20-DoF dexterous hand in real time. Formally, let $p_t\in\mathbb{R}^{15}$ be the concatenated fingertip positions and $q_t\in\mathbb{R}^{20}$ be the robot hand joint angles. We learn a retargeting function $f_\theta:\mathbb{R}^{15}\rightarrow\mathbb{R}^{20}$, and at runtime we directly set $q_{hand}=f_\theta(p_t)$, which is concatenated with $q_{body}$ to form the full reference command for the low-level controller.

We formulate this as a regression problem to achieve constant-time inference and low operational overhead. Specifically, we parameterize $f_\theta$ as a lightweight MLP that maps fingertip positions to robot joint angles, and train it on a paired dataset $\mathcal{D}={(p_t,q_t)}$ using a mean-squared error (MSE) objective:
\begin{equation}
\min_{\theta} \ \mathbb{E}{(p,q)\sim \mathcal{D}}\big[|f{\theta}(p) - q|_2^2\big].
\end{equation}

The training dataset $\mathcal{D}$ is generated via an offline optimization-based retargeting process~\cite{wuji2025retargeting} across a diverse range of hand poses. Once trained, the MLP $f_\theta$ generalizes to various hand sizes and provides smooth, continuous joint trajectories without the high computational cost of per-frame optimization.

\subsection{Two-Stage Imitation Learning with Human Data} \label{sec:method_learning}

\paragraph{Policy Learning Setup}
We perform imitation learning with the following setup. We use Action Chunking Transformer (ACT)~\cite{zhao2023learningfinegrainedbimanualmanipulation} as our policy backbone with a ResNet-18 pre-trained visual encoder. The observation space consists of an RGB image $I_t$ of shape 640$\times$480 from the G1 built-in RealSense D435i camera, and the proprioceptive state $s_t$ comprising joint positions of the body and hands. The action space $a_t$ includes 20 dimensions per hand and 35 dimensions for the humanoid body. Following TWIST2~\cite{ze2025twist2}, we predict low-level motion tracker targets instead of direct joint positions, enabling real-world adjustment and balancing.

\paragraph{Two-Stage Training with Human Data}
Our tracking system enables collecting human demonstration data more efficiently than robot teleoperation. However, robot proprioceptive states $\mathbf{s}_t$ are not directly available from human data. We approximate them using the previous timestep's action $\mathbf{a}_{t-1}$, validated by analyzing action-state correspondence in robot data, which confirms $\mathbf{a}_t$ corresponds to $\mathbf{s}_{t+1}$ with minimal latency.

Due to morphological differences between humans and the G1 robot, human motion cannot be directly replayed. Rather than aligning human and robot data for mixed training~\cite{kareer2025egomimic}, we train sequentially: stage 1 only on human data to learn generalizable motion priors, stage 2 on robot data to refine toward robot embodiment. This enables fast human data collection without alignment.

\begin{table*}[t]
\centering

\resizebox{0.9\textwidth}{!}{%
\begin{tabular}{llcccccc}
\toprule
\multirow{2}{*}{Metric} & \multirow{2}{*}{Method} & \multicolumn{6}{c}{Tasks} \\
\cmidrule(l){3-8} 
 & & \textit{Scan\&Pack} & \textit{Hang Towel} & \textit{Open Door} & \textit{Place Basket} & \textit{Pick Bread} & \textbf{Avg$^\dag$} \\
\midrule

\multirow{2}{*}{Collection Time (min) $\downarrow$} 
 & Baseline & - & 68 & 75 & 46 & 50 & 59.8\\
 & Ours & \textbf{78} & \textbf{53} & \textbf{62} & \textbf{35} & \textbf{27} & \textbf{44.3}\\
\midrule

\multirow{2}{*}{Collection Success Rate $\uparrow$} 
 & Baseline & 0/60 & 26/60 & 52/60 & 47/60 & 54/60 & 74.6\%\\
 & Ours & \textbf{54/60} & \textbf{50/60} & \textbf{57/60} & \textbf{58/60} & \textbf{55/60} & \textbf{91.7\%}\\
\midrule

\multirow{2}{*}{Policy Success Rate $\uparrow$} 
 & Baseline & 0/30 & 11/30 & 10/30 & 22/30 & 26/30 & 57.5\%\\
 & Ours & \textbf{20/30} & \textbf{19/30} & \textbf{22/30} & \textbf{26/30} & \textbf{29/30} & \textbf{80.0\%}\\
\bottomrule
\end{tabular}
}
\caption{\textbf{Data Collection Comparison.} We report the total time to collect 60 episodes, the number of successful teleoperation attempts out of 60, and the success rate of the trained policy over 30 trials. (-) indicates the task was infeasible with the baseline setup. 
\textbf{Note:} All values in the \textbf{Avg} column are calculated on the subset of four tasks successfully performed by both methods (excluding \textit{Scan\&Pack}).}
\label{tab:data_collection}
\end{table*}

\section{Experiments}

With our experiments, we seek to answer the following questions: (A) How does our system's data collection efficiency and quality compare to existing methods on challenging whole-body dexterous tasks? (B) How to evaluate our system's capability of hand retargeting, specifically for the dynamic tasks. (C) Does incorporating human data improve generalization to new positions, objects, and backgrounds?

\subsection{Whole-Body Dexterous Teleoperation}

\subsubsection{Tasks} We evaluate teleoperation success rate and per-episode time on a suite of challenging loco-manipulation tasks, as visualized in Fig.~\ref{fig:tasks}.

\smallskip

\noindent \textbf{\textit{\underline{Scan\&Pack}}}

\smallskip
    
\textbf{Goal and Success Criteria.} The robot grasps a barcode scanner with one hand and a toy with the other, then pulls the trigger to scan the barcode on the toy. It then places the toy into a shopping bag, grasps the bag handle, and hands it to a human. A successful trial requires: the scanner trigger pulled with the scanner aimed at the barcode (indicated by red light), the toy inside the shopping bag, and the robot lifting the bag from the table.
            
\textbf{Main Challenge.} The scanning step requires \textbf{high dexterity}, as the robot must stably hold the scanner while using a spare finger to pull the trigger. This makes the task extremely challenging for simpler hand hardware used in previous works. Additionally, this task is \textbf{long-horizon}.

\smallskip
\noindent \textbf{\textit{\underline{HangTowel}}}
\smallskip
    
\textbf{Goal and Success Criteria.} The robot grasps a hanger from a rod, threads a towel through the hanger's horizontal bar with the other hand, and returns the hanger to the rod. Success requires the towel draped over the bar and the hanger back on the rod.
            
\textbf{Main Challenge.} Successful completion requires \textbf{bimanual coordination} when threading the towel onto the hanger, as well as {robustness to disturbance}, since the hanger may rotate upon contact when reaching and grasping.

\smallskip 

\noindent \textbf{\textit{\underline{OpenDoor}}}

\smallskip 
    
\textbf{Goal and Success Criteria.} The robot needs to grasp a real-world office door handle, press it down to unlatch the door, and walk forward to push the door open. The task is considered successful if the door is opened more than 45 degrees.
            
\textbf{Main Challenge.} This task involves \textbf{articulated object manipulation}, as the robot must grasp and press down the handle to unlatch before pushing. It also requires \textbf{coordinated locomotion}, since the robot needs to walk forward while pushing the door open.

\smallskip 

\noindent \textbf{\textit{\underline{PlaceBasketOnShelf}}}
\smallskip
    
\textbf{Goal and Success Criteria.} The robot squats down to pick up a basket, stands up, rotates its upper body, and places the basket on a small shelf. The task is considered successful if the basket ends up stably on the shelf.
            
\textbf{Main Challenge.} This task requires coordinated \textbf{whole-body movement} to be completed.

\smallskip
    
        

\noindent \textbf{\textit{\underline{PickBread}}}
\smallskip

\textbf{Goal and Success Criteria.} The robot grasps a bread roll from the table, places it into a basket, and returns. The task is considered successful if the bread is stably placed inside the basket.

\textbf{Main Challenge.} In generalization evaluation, this task requires grasping of objects with diverse geometry.

\subsubsection{Baseline and Comparison}

\textbf{Baseline.} We compare our system with existing vision-based teleoperation systems. In our baseline, the humanoid motion is teleoperated with PICO following the setup in TWIST2. Yet in the original TWIST2 controller, triggers are used to control the binary open-close of the hand, so we further add the vision-based hand tracking module to allow dexterous hand teleoperation. 

\textbf{Setup and Metrics.} For each task and teleoperation solution, we perform 60 teleoperation attempts (operators practice in advance). We compute the successfully teleoperated trails in these attempts, and the total time used. To compare the data quality of obtained demonstrations, we train an imitation learning policy on collected demos following the setup in~\ref{sec:method_learning} (no human data). We compare the task success rate of the trained policies and report them in Table ~\ref{tab:data_collection}.

\subsubsection{Results}

The quantitative results of our whole-body teleoperation evaluation are summarized in Table~\ref{tab:data_collection}.
First, we note that the baseline system was unable to perform the \textit{Scan\&Pack} task due to severe occlusion issues, whereas our system achieved a 90\% teleoperation success rate.
To ensure a fair quantitative comparison, the following average metrics are calculated only on the four tasks feasible for both systems (\textit{Hang Towel, Open Door, Place Basket, Pick Bread}).

\textbf{Task Coverage and Reliability.}
The baseline's failure in \textit{Scan\&Pack} highlights the fundamental limitations of vision-based teleoperation: susceptibility to jitter and severe occlusion.
In this task, when the operator grasps the barcode scanner, the device itself occludes a significant portion of the hand from the headset's cameras.
For vision-based algorithms, this loss of visual features leads to tracking failure or excessive jitter, making it impossible to precisely control the index finger to pull the trigger.
This failure mode is critical because tool use—which inherently causes self-occlusion—is ubiquitous in real-world daily tasks.
Consequently, vision-based methods are often restricted to simple, non-occluded interactions where the full hand remains visible at all times.
In contrast, \emph{HumDex} is immune to visual occlusion, reliably capturing fine-grained finger articulation during complex tool manipulation, achieving a 90\% teleoperation success rate on this challenging task.

\textbf{Collection Efficiency.}
On the common task set, \emph{HumDex} demonstrates significantly higher efficiency. The average time required to collect 60 episodes is reduced from 59.8 minutes (Baseline) to 44.3 minutes (Ours), representing a \textbf{26\% improvement}.
This efficiency gain is largely attributed to the robustness of our IMU-based tracking against occlusion constraints.
In the baseline vision-based system, the operator is strictly required to keep their hands within the headset's field of view to prevent tracking loss. This constraint forces the operator to frequently adjust their head orientation to follow their hands, introducing unnatural body movements that compromise the stability of the data collection process. Furthermore, inevitable momentary tracking losses often necessitate pausing or resetting the episode to regain lock.
In contrast, our system allows for unconstrained motion, enabling the operator to focus entirely on task execution without managing tracking visibility, resulting in a smoother and faster workflow.

\textbf{Success Rate and Quality.}
Our system also achieves higher reliability. The average teleoperation success rate on the common tasks is \textbf{91.7\%} for \emph{HumDex}, compared to \textbf{74.6\%} for the baseline.
Crucially, this improvement in data quality translates to downstream policy performance. Policies trained on our data achieve an average success rate of \textbf{80.0\%}, significantly outperforming the baseline's \textbf{57.5\%}.
For challenging bimanual tasks like \textit{Hang Towel}, the performance gap is particularly evident (19/30 vs. 11/30), confirming that our system captures the precise, smooth motions required for complex manipulation.

\subsection{Dexterous Hand Retargeting}
\label{sec:exp_hand}

We evaluate the effectiveness of our learning based hand retargeting on the Wuji dexterous hand using the Vdhand. We compare against a classical optimization-based retargeting baseline that solves a constrained inverse-kinematics / nonlinear least-squares problem per frame. Both methods consume the same human hand observations and output 20-DoF Wuji hand joint targets at the same control rate.

\textbf{Input/Output.} The retargeting input is a compact 15D representation formed by the 3D positions of five fingertips (thumb, index, middle, ring, little) in the glove hand frame. The output is the 20-DoF joint angle vector of the Wuji hand. 

\begin{figure}[t]
    \centering
    \includegraphics[width=\linewidth]{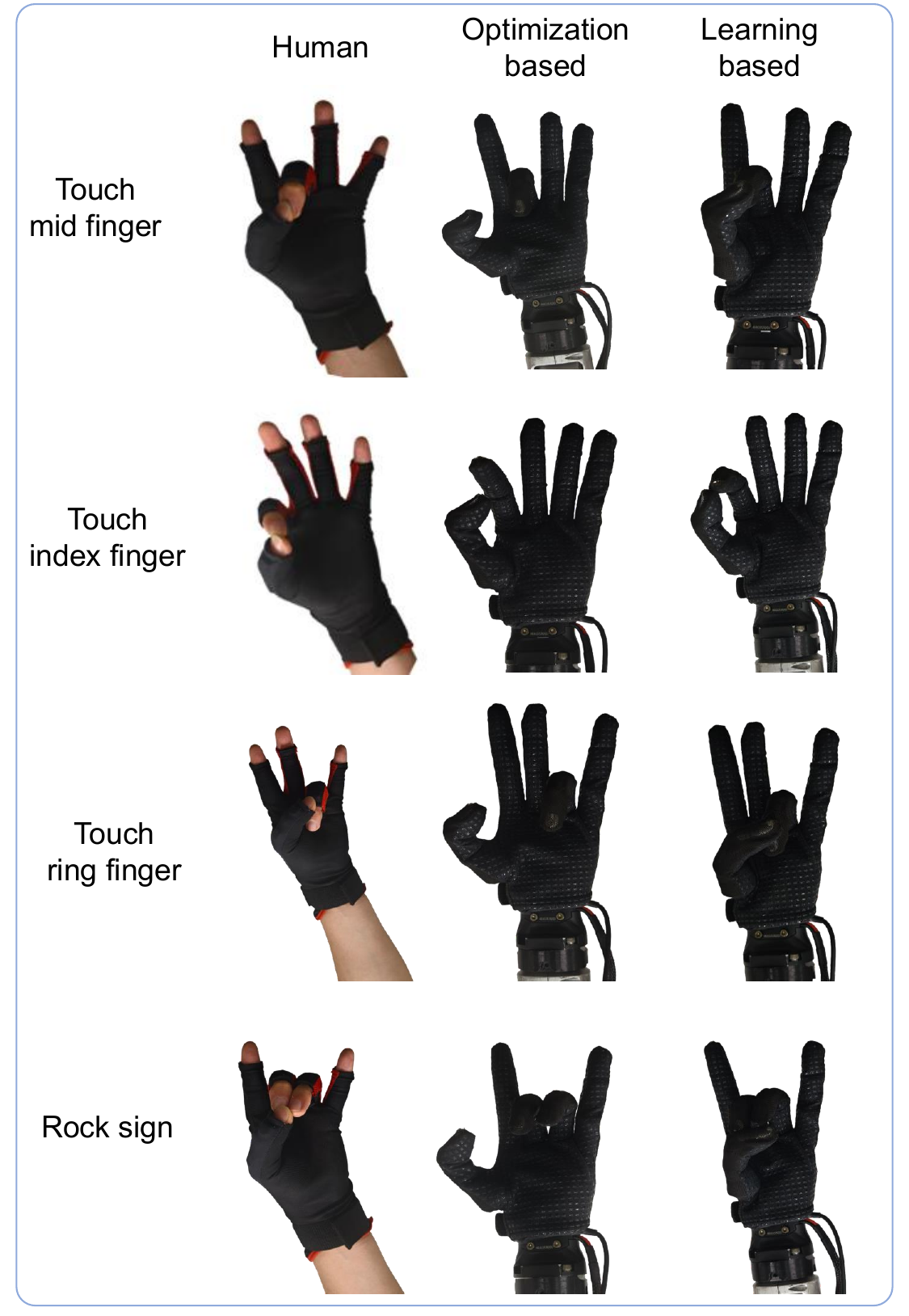}
    \caption{\textbf{Qualitative pose reproduction on the Wuji hand.}
    We compare an optimization-based retargeting baseline and our learning-based retargeter on canonical dexterous poses captured by the inertial glove, including \textit{touch middle finger}, \textit{touch index finger}, \textit{touch ring finger}, and the \textit{rock sign}.}
    \label{fig:qual_hand_poses}
\end{figure}

\textbf{(B1) Qualitative motion/pose reproduction.}
We first evaluate whether a retargeting method can reproduce representative dexterous hand poses \emph{faithfully and stably} under a controlled open-loop replay setting. This evaluation intentionally removes whole-body locomotion and task interaction, so that observed differences can be attributed primarily to the hand retargeting module itself. Concretely, we curate a set of canonical poses that stress finger individuation and cross-finger coordination, including \textit{touch middle finger}, \textit{touch index finger}, \textit{touch ring finger}, \textit{one-finger extension}, and the \textit{rock sign} (Fig.~\ref{fig:qual_hand_poses}). For each pose, we feed the same glove-tracked input---the 3D positions of the five fingertips---to each retargeting method, and replay the resulting joint trajectories on the Wuji hand. We then present side-by-side snapshots comparing our learning-based retargeter against an optimization-based baseline.

Our qualitative assessment focuses on two observable criteria: 
(1) \emph{pose correctness}, i.e., whether the intended finger configuration is achieved (e.g., the specified finger reaches/contacts the thumb while non-target fingers remain in the desired state);
(2) \emph{unwanted coupling}, i.e., whether irrelevant fingers exhibit spurious bending, participation, or collapse to easier but incorrect configurations; and 
This qualitative study provides an interpretable sanity check of hand mapping quality and demonstrates that our learned retargeter yields more reliable pose realization than per-frame optimization in representative dexterous configurations.

\textbf{(B2) Teleoperation task evaluation.}
We further evaluate retargeting performance under closed-loop humanoid teleoperation with \emph{task-critical hand sub-tasks} that impose the highest dexterity requirements in our data-collection suite, so that grasp stability and contact transitions directly determine success. Concretely, we derive three evaluation sub-tasks from the full demonstrations: \textit{Scanner Triggering} (from \textit{Scan\&Pack}), \textit{Hanger Stabilization} (from \textit{HangTowel}), and \textit{Doll Grasping} (from \textit{Scan\&Pack}). Importantly, we enforce stricter success criteria to better reflect real-world use. 
In \textit{Scanner Triggering}, the robot must actuate the scanner trigger \emph{five consecutive times}, and each actuation must be \emph{held for 3 seconds}, matching realistic scanning behavior rather than a single tap. 
In \textit{Hanger Stabilization}, the robot must maintain a stable grasp on the hanger while the operator rotates the wrist \emph{back and forth for five cycles}, approximating the disturbance and wrist motion that occur when threading a towel; a trial is successful only if the hanger remains securely grasped throughout the entire sequence. 
In \textit{Doll Grasping}, we increase difficulty by using a doll that is \emph{3$\times$ larger} than the one in the original task, making it harder to envelop and stabilize. The results are shown in (Tab.~\ref{tab:hand_success_3x4})

\textbf{Efficiency.}
Besides effectiveness, we emphasize deployment efficiency. Our retargeting model is a lightweight MLP trained on paired samples of five-fingertip positions and robot joint targets, requiring only a short calibration dataset(typically about 20k frames, i.e., and less than 20 minutes of recording).

\begin{table}[t]
\centering
\setlength{\tabcolsep}{6pt}
\renewcommand{\arraystretch}{1.15}
\begin{tabular}{lcccc}
\toprule
Task & \makecell{\textit{Glove}+\\\textit{Learn.}}
     & \makecell{\textit{Glove}+\\\textit{Opt.}}
     & \makecell{\textit{PICO}+\\\textit{Learn.}}
     & \makecell{\textit{PICO}+\\\textit{Opt.}} \\
\midrule
\textit{Scanner Triggering} & 20/30 & 14/30 & 13/30 & 12/30 \\
\textit{Hanger Stabilization}  & 20/30 & 21/30 & 15/30 & 18/30 \\
\textit{Doll Grasping}  & 29/30 & 23/30 & 17/30 & 15/30 \\
\bottomrule
\end{tabular}
\caption{Teleoperation success rate (\#success/\#trials) across tasks under different hand retargeting interfaces.}
\label{tab:hand_success_3x4}
\end{table}

\subsection{Human Data for Policy Generalization}
\label{sec:exp_human_data}

We evaluate how leveraging human data improves policy generalization on the \textit{PickBread} task.

\textbf{Data Collection.} We use our portable system to efficiently collect a dataset covering diverse environmental variations. Specifically, the dataset consists of:
\begin{itemize}
    \item \textbf{Robot Data (Base):} 50 episodes of teleoperation data where the robot picks up a bread roll from a fixed position on a plain table.
    \item \textbf{Human Data (Position Var.):} 100 episodes of human operation where the bread is placed at random positions across the table surface.
    \item \textbf{Human Data (Object Var.):} 300 episodes (3 items $\times$ 100 each) of human operation picking diverse items (apple, banana, leaf) to introduce visual and shape diversity.
    \item \textbf{Human Data (Background Var.):} 300 episodes (3 backgrounds $\times$ 100 each) of human operation picking the bread with varied table textures and distractors.
\end{itemize}

\textbf{Comparison Baselines.} We compare two primary policies: (1) \textit{RobotOnly}, trained solely on the 50 robot episodes; and (2) \textit{Ours}, trained using our two-stage framework (human pre-training $\rightarrow$ robot fine-tuning). 
To validate our sequential design, we also attempted to train a policy by naively mixing human and robot data (\textit{Mix}). However, this baseline failed to converge and achieved a \textbf{0\% success rate}, likely due to the conflicting gradient signals arising from the significant embodiment gap between human and robot kinematics.

\textbf{Generalization Evaluation.} To demonstrate that our framework learns robust representations, we evaluate success rates across three out-of-distribution (OOD) settings:
\begin{enumerate}
    \item \textit{UnseenPos:} The object is placed in locations not covered in the robot dataset (but within the range of human data).
    \item \textit{UnseenObj:} The target object is replaced with new items unseen in robot data (but seen in human data).
    \item \textit{UnseenBg:} The background is altered with new tablecloths unseen in robot data (but seen in human data).
\end{enumerate}

\textbf{Results.}
The quantitative results are reported in Table~\ref{tab:human_generalize}. Each policy is evaluated for 30 trials under each setting.

\begin{table}[h]
\centering
\resizebox{\columnwidth}{!}{%
\begin{tabular}{lcccc}
\toprule
\textbf{Settings} & \textbf{Seen} & \makecell{\textbf{Unseen}\\\textbf{Position}} & \makecell{\textbf{Unseen}\\\textbf{Object}} & \makecell{\textbf{Unseen}\\\textbf{Background}}\\
\midrule
Robot Data Only & 29/30 & 12/30 & 10/30 & 9/30 \\
\textbf{Ours} & \textbf{30/30} & \textbf{21/30} & \textbf{20/30} & \textbf{25/30} \\
\bottomrule
\end{tabular}
}
\caption{Policy Generalization Success Rate (\%). We compare the policy trained only on robot data against our two-stage approach leveraging human data.}
\label{tab:human_generalize}
\end{table}

\textbf{Human data improves generalization at a lower cost.} 
As illustrated in Table~\ref{tab:human_generalize}, the \textit{RobotOnly} policy suffers from severe overfitting. While it performs perfectly in the "Seen" setting, its performance drops drastically (e.g., to 30\% in UnseenBg) when facing distribution shifts. 
In contrast, \textit{Ours} significantly boosts robustness, improving success rates by nearly \textbf{2$\times$ across all generalization settings}. This confirms that pre-training on diverse human data allows the policy to learn invariant visual features and high-level motion priors—such as "reach towards the object regardless of background"—which are successfully transferred to the robot embodiment during fine-tuning.

\textbf{Sequential training is essential for bridging the gap.} 
The complete failure of the \textit{Mix} baseline (0\% success) highlights the challenge of the embodiment gap. Direct mixing forces the network to map similar visual states to two distinct action spaces (human vs. robot), causing optimization conflicts. Our sequential approach resolves this by using human data to learn generalizable features first, and then using robot data solely to adapt the action output to the robot's specific kinematics, effectively combining the diversity of human data with the precision of robot data.

\section{Conclusion}
We presented \ours, a portable teleoperation system for humanoid whole-body dexterous manipulation. By combining IMU-based full-body motion capture with a learning-based dexterous hand retargeting module, \ours~ enables reliable demonstration collection for long-horizon, contact-rich manipulation tasks that are difficult or infeasible for vision-based or infrastructure-heavy systems. Our system significantly improves data collection efficiency, teleoperation success rate, and downstream policy performance across a diverse set of whole-body manipulation tasks.

Beyond teleoperation, \ours~ enables efficient collection of human whole-body motion data in unconstrained environments. We show that, when leveraged through a two-stage imitation learning framework, such human data provides strong motion and perception priors that substantially improve generalization to unseen object positions, object categories, and visual backgrounds---while requiring only a small amount of robot-specific data for embodiment adaptation. Together, these results suggest that portable human motion capture, combined with appropriate training strategies, offers a practical path toward scalable data acquisition for general-purpose humanoid manipulation. We plan to open-source the full system to support future research in humanoid dexterous manipulation.

\section{Limitation.} 
Although \ours~ achieves strong whole-body dexterous manipulation capabilities, several challenges remain. First, due to computational and time constraints, we are unable to scale the training data to a larger scale, which may further improve performance. Second, while the collected hand data is sufficient for the tasks studied, extending to a broader range of hand postures, contact modes, and force-sensitive interactions remains an open challenge. Third, hardware payload limits and actuation strength prevent us from exploring potentially more capable manipulation behaviors. We leave addressing these limitations to future work.

\newpage

\bibliographystyle{plainnat}
\bibliography{references}

\clearpage
\appendix

\setcounter{figure}{0}
\setcounter{table}{0}
\renewcommand{\thefigure}{A\arabic{figure}}
\renewcommand{\thetable}{A\arabic{table}}

\section*{Method Additional Details}
\label{sec:supp_method}

\subsection{Hardware Setup}
\label{sec:supp_hardware}

We visualize our hardware integration and teleoperation setup in Fig.~\ref{fig:supp_hardware}. Our system is composed of three key components:

\begin{figure}[h]
    \centering
    \includegraphics[width=\linewidth]{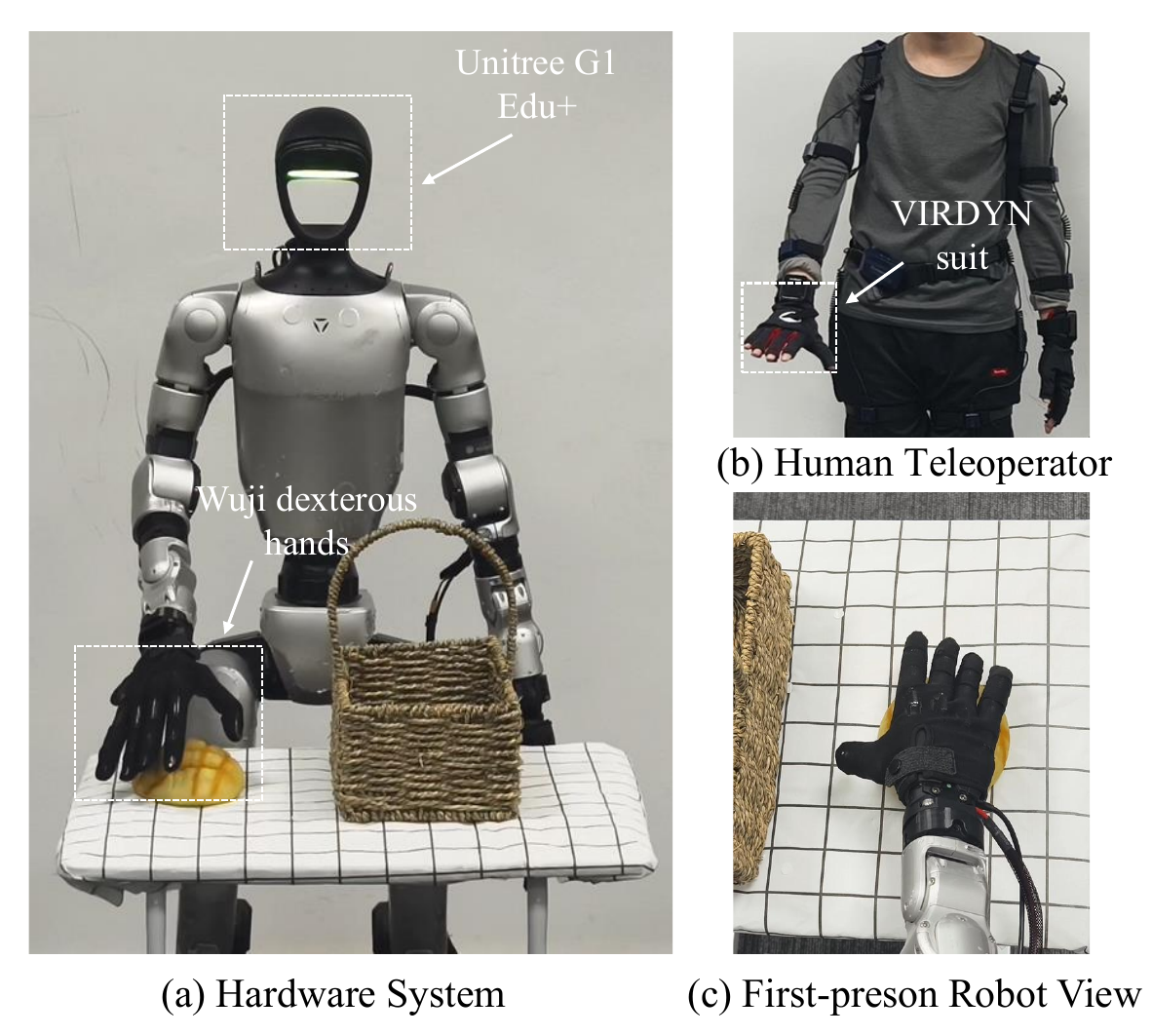}
    \caption{\textbf{System Hardware Overview.} \textbf{(a)} The Unitree G1 Edu+ humanoid robot integrated with custom WUJI dexterous hands. \textbf{(b)} The human operator wearing the VIRDYN inertial motion capture suit and data gloves for immersive teleoperation. \textbf{(c)} Visual feedback is provided by the robot's built-in RealSense camera. }
    \label{fig:supp_hardware}
\end{figure}

\begin{itemize}
    \item \textbf{Robot Platform:} We utilize the \textbf{Unitree G1 Edu+} humanoid robot. This upgraded version features \textbf{29 active degrees of freedom (DoF)} to support complex whole-body locomotion and manipulation. For visual perception, we utilize the robot's built-in Intel RealSense D435i camera to capture high-fidelity egocentric RGB observations.
    
    \item \textbf{Dexterous Hands:} The robot is equipped with two custom \textbf{WUJI} dexterous hands. Each hand possesses \textbf{20 actuated DoFs}, featuring 4 independent DoFs per finger. This high-dimensional actuation space enables the execution of fine-grained manipulation tasks comparable to human hand dexterity.
    
    \item \textbf{Teleoperation Interface:} A key advantage of our framework is its highly modular hardware design, allowing for the free combination of different body and hand tracking solutions. To ensure both accessibility and precision, we decouple the hardware into two independent modules:
    \begin{itemize}
        \item \textit{Body Tracking Options:} We provide two distinct full-body tracking configurations. The primary setup utilizes the \textbf{Vdmocap} 15-node commercial suit for high-precision baseline tracking on standard kinematic chains. As a highly cost-effective (\textless \$200) alternative, we also custom-built a scalable system based on the open-source \textbf{SlimeVR} ecosystem. It employs ICM-45686 6-axis IMUs and nRF52840 SoCs (running the ESB protocol) to ensure stable, low-latency connectivity in complex Wi-Fi environments.
        
        \item \textit{Dexterous Hand Tracking Options:} For fine-grained manipulation, the system integrates seamlessly with commercial high-fidelity inertial gloves, specifically supporting both \textbf{Vdhand} and \textbf{Manus}. Each glove utilizes an array of IMUs distributed across the dorsum and finger segments. This purely inertial approach ensures that complex grasping motions are accurately captured even under severe self-occlusion.
    \end{itemize}
    Crucially, our control pipeline is hardware-agnostic. Operators can freely mix and match these body and hand modules without requiring any modifications to the underlying retargeting algorithms.
\end{itemize}

\subsection{Hand Retargeting Details}
\textbf{Data Collection.} To train the retargeting function, we collected a paired dataset of human hand poses and robot hand joint configurations.
The operator wears the data glove and performs a predefined set of random finger movements and canonical poses. The corresponding robot joint angles are solved via an offline optimization-based IK solver to ensure kinematic feasibility. The collection process takes approximately 20 minutes, resulting in $\sim$20k paired frames.

\textbf{Training Hyperparameters.}
The retargeting network is a Multi-Layer Perceptron (MLP).
\begin{itemize}
    \item \textbf{Architecture:} The IK model employs a finger-wise modular design. For each of the five fingers, the sub network is structured as: [Input(3) $\rightarrow$ FC(128) $\rightarrow$ LeakyReLU $\rightarrow$ BN $\rightarrow$ FC(128) $\rightarrow$ LeakyReLU $\rightarrow$ BN $\rightarrow$ FC(4) $\rightarrow$ Tanh]. The outputs from all sub-networks are concatenated to form the final Output(20).
    \item \textbf{Optimizer:} AdamW ($lr=1e-4$, $\beta_1=0.9, \beta_2=0.999$).
    \item \textbf{Batch Size:} 2048.
    \item \textbf{Epochs:} 300 (with early stopping based on validation loss).
\end{itemize}

\subsection{Data Collection Process}
We describe the specific protocols for the two data sources. All data is managed via a Redis-based distributed framework to ensure low-latency synchronization between vision and proprioception.

\textbf{1. Human Data Collection}
\begin{itemize}
    \item \textbf{Setup \& Vision:} The operator wears the VIRDYN suit and gloves without active robot execution. To emulate the robot's egocentric view, an Intel RealSense D435i camera is mounted on a neckband worn by the operator.
    \item \textbf{Pipeline:} Raw motion capture data is processed via the General Motion Retargeting (GMR) model. This optimization-based solver maps human motion to the robot's 35-DoF configuration space in real-time.
    \item \textbf{Recorded Modalities:} We record the egocentric RGB images (640$\times$480 resolution) alongside the retargeted 35-DoF whole-body targets (output of GMR) and 40-DoF bimanual hand targets (20 DoF per hand).
    \item \textbf{Frequency \& Trigger:} The teleoperation and retargeting loop runs at 100 Hz for smoothness, while the recorder saves data at 30 Hz. A USB foot pedal is used to trigger episode start/stop for clean segmentation.
\end{itemize}

\textbf{2. Robot Data Collection}
\begin{itemize}
    \item \textbf{Setup \& Vision:} The operator controls the physical G1 robot via the teleoperation system. We utilize the robot's built-in Intel RealSense D435i head camera for vision.
    \item \textbf{Pipeline:} The operator's motion is mapped to robot commands via the same GMR solver in real-time. These commands are executed by the robot and simultaneously fetched from Redis for recording, synchronized with the ZMQ video stream.
    \item \textbf{Recorded Modalities:} We record the egocentric RGB images (640$\times$480 resolution) alongside the full robot state (23-DoF body + 40-DoF hand joints) and the applied action commands (35-DoF whole-body targets + 40-DoF bimanual hand targets).
    \item \textbf{Frequency \& Trigger:} The recording script runs at 30 Hz. Similar to the human phase, the same global foot pedal trigger is used to manage episode recording and save data in a standardized JSON + JPEG format.
\end{itemize}

\subsection{Policy Learning Details}

\textbf{Input/Output Modalities.}
\begin{itemize}
    \item \textbf{Observation Space:} Includes egocentric RGB images ($480 \times 640$) and robot joint positions (Whole Body 31 dimensions + 20 dimensions per hand).
    \item \textbf{Action Space:} Target joint positions for the whole body (including 35 DoF for body and 20 DoF per hand).
\end{itemize}

\textbf{Network and Training.}
We employ a standard Action Chunking Transformer~\cite{zhao2023learningfinegrainedbimanualmanipulation} backbone.
\begin{itemize}
    \item \textbf{Hardware:} All models are trained on a single NVIDIA RTX 4090 GPU.
    \item \textbf{Single-stage training on robot data:} 8000 epochs for \textit{HangTowel}, \textit{Scan\&Pick}, \textit{OpenDoor} and \textit{PlaceBasketOnShelf}, 5000 epochs for \textit{PickBread}. Initial learning rate $2e^{-5}$. Batch size 16. Chunk size 200 for  \textit{PickBread}, 400 for \textit{OpenDoor} and 300 for the rest. Takes 2$\sim$7 hours depending on demonstration trajectory length and number of episodes. 
    \item \textbf{Two-stage training with human data:} Stage 1 trained on human data only for 1000 epochs ($\sim$5 hours). Batch size 16. Stage 2 trained on robot data only for 4000 epochs ($\sim$ 3 hours). Batch size 16.
    \item \textbf{Inference:} During deployment, the policy inference runs at 30 Hz.
\end{itemize}

\section*{Additional Experiment Details}
\label{sec:supp_exp}

\subsection{Teleoperation System Validation}
\label{sec:teleop_val}

To assess the precision and reliability of our low-cost teleoperation solution, we conducted a comparative study against the commercial Vdmocap system. We also performed an ablation study on the number of trackers within the SlimeVR setup (14, 10, and 7 nodes).
We selected three representative tasks from those detailed in the main paper—\textit{HangTowel}, \textit{PlaceBasketOnShelf}, and \textit{PickBread}—to evaluate performance across different manipulation complexities.

As shown in Table~\ref{tab:teleop_success}, the results indicate that:
\begin{itemize}
    \item \textbf{Commercial vs. Low-Cost:} The 14-node SlimeVR setup achieves a success rate comparable to the commercial baseline (e.g., 48/60 vs. 50/60 on \textit{HangTowel}), proving that our $<\$200$ solution is sufficient for high-quality data collection.
    \item \textbf{Tracker Density:} Reducing tracker count degrades performance, particularly in whole-body tasks. While the 10-node setup remains robust for simpler reaching (\textit{PlaceBasket}), the 7-node configuration suffers a significant performance drop (42/60 on \textit{HangTowel}), highlighting the necessity of the full 14-node configuration for complex, coordinated manipulation.
\end{itemize}

\begin{table}[h]
    \centering
    \caption{\textbf{Teleoperation Success Rate Comparison.} We report success trials out of 60 attempts. The 14-node SlimeVR setup yields performance close to the commercial VIRDYN suit, whereas reducing nodes (7 or 10) leads to lower reliability in complex tasks.}
    \label{tab:teleop_success}
    \resizebox{\linewidth}{!}{
    \begin{tabular}{l c c c c}
        \toprule
        \multirow{2}{*}{\textbf{Task}} & \multirow{2}{*}{\textbf{VIRDYN}} & \multicolumn{3}{c}{\textbf{SlimeVR Configurations}} \\
        \cmidrule(l){3-5}
         & & \textbf{14 Nodes} & \textbf{10 Nodes} & \textbf{7 Nodes} \\
        \midrule
        \textit{HangTowel}   & 50/60 & 48/60 & 44/60 & 42/60 \\
        \textit{PlaceBasket} & 58/60 & 57/60 & 57/60 & 51/60 \\
        \textit{PickBread}   & 55/60 & 55/60 & 54/60 & 53/60 \\
        \bottomrule
    \end{tabular}
    }
\end{table}

\subsection{Failure Case Analysis}
\label{sec:failure_analysis}

\begin{figure*}[h]
    \centering
    \includegraphics[width=0.8    \linewidth]{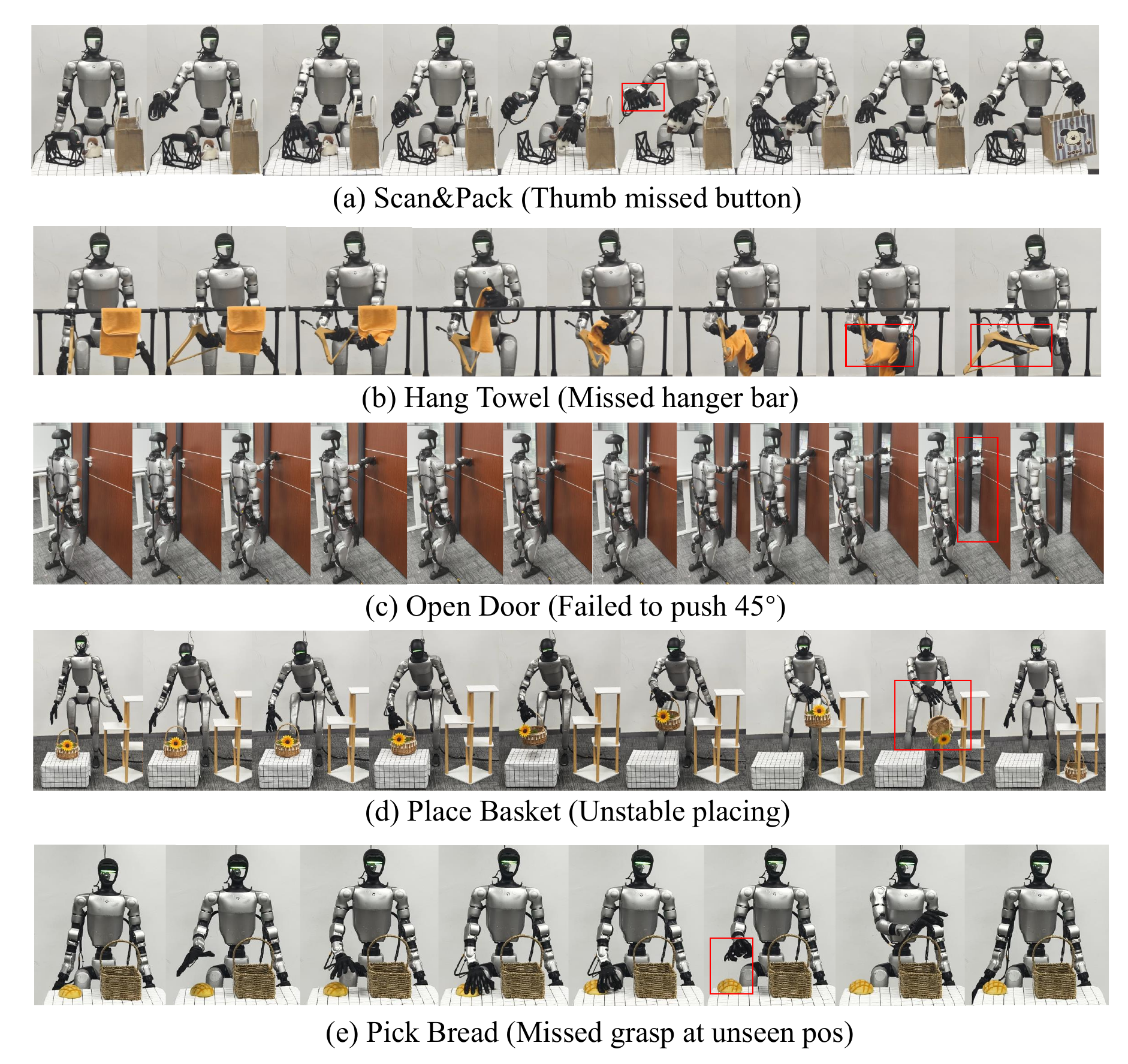}
    \caption{\textbf{Visualization of Failure Cases.} We showcase common failure modes for each task: 
    (a) \textbf{Scan\&Pack:} The thumb fails to press the trigger due to an inappropriate initial grasp. 
    (b) \textbf{Hang Towel:} The towel misses the hanger due to misalignment between the two hands. 
    (c) \textbf{Open Door:} The robot unlatches the door but fails to open it fully ($>45^{\circ}$) due to insufficient forward locomotion. 
    (d) \textbf{Place Basket:} The basket is not placed stably due to whole-body uncoordination and foot instability. 
    (e) \textbf{Pick Bread:} The robot grasps empty air due to estimation errors when the bread is placed in an unseen generalized position.}
    \label{fig:failure_cases}
\end{figure*}

We provide a qualitative analysis of typical failure modes observed across the five tasks. Representative failure sequences are visualized in Fig.~\ref{fig:failure_cases}.

Detailed analysis for each task is as follows:

\begin{itemize}
    \item \textbf{Scan \& Pack (Dexterous Manipulation Failure):} 
    As shown in the first row of Fig.~\ref{fig:failure_cases}, the primary failure mode is the inability to trigger the scanner. This typically occurs when the initial grasp on the scanner handle is too low or rotated. Although the robot holds the object, the thumb's kinematic reach is restricted, preventing it from pressing the button effectively.
    
    \item \textbf{Hang Towel (Bimanual Coordination Failure):} 
    Failures in this task often stem from spatial misalignment between the left hand (holding the hanger) and the right hand (holding the towel). If the coordination policy fails to synchronize the two end-effectors precisely, the towel may hit the rim of the hanger and fall, rather than threading through the bar.
    
    \item \textbf{Open Door (Loco-Manipulation Failure):} 
    The crucial failure mode involves the coordination between the arm and the base. In some cases, the robot successfully unlatches the door handle, but the locomotion policy fails to execute a continuous forward walk. Consequently, the door is merely unlatched or slightly ajar but does not reach the required $45^{\circ}$ opening angle.
    
    \item \textbf{Place Basket (Whole-Body Stability Failure):} 
    This task requires significant vertical motion (squatting and standing). Failures occur when the robot loses balance or exhibits jittery foot movement during the lifting phase. This instability propagates to the upper body, causing the basket to be placed on the edge of the shelf and subsequently fall off.
    
    \item \textbf{Pick Bread (Generalization Failure):} 
    When evaluating on unseen positions (Generalization Setting), the policy occasionally fails to adapt to large spatial shifts. As illustrated in the last row, the robot may execute a grasp at a position offset from the actual bread, resulting in a grasp on empty air. This indicates the limitations of the policy's spatial generalization boundary.
\end{itemize}

\begin{figure*}[h]
    \centering
    \includegraphics[width=0.7\linewidth]{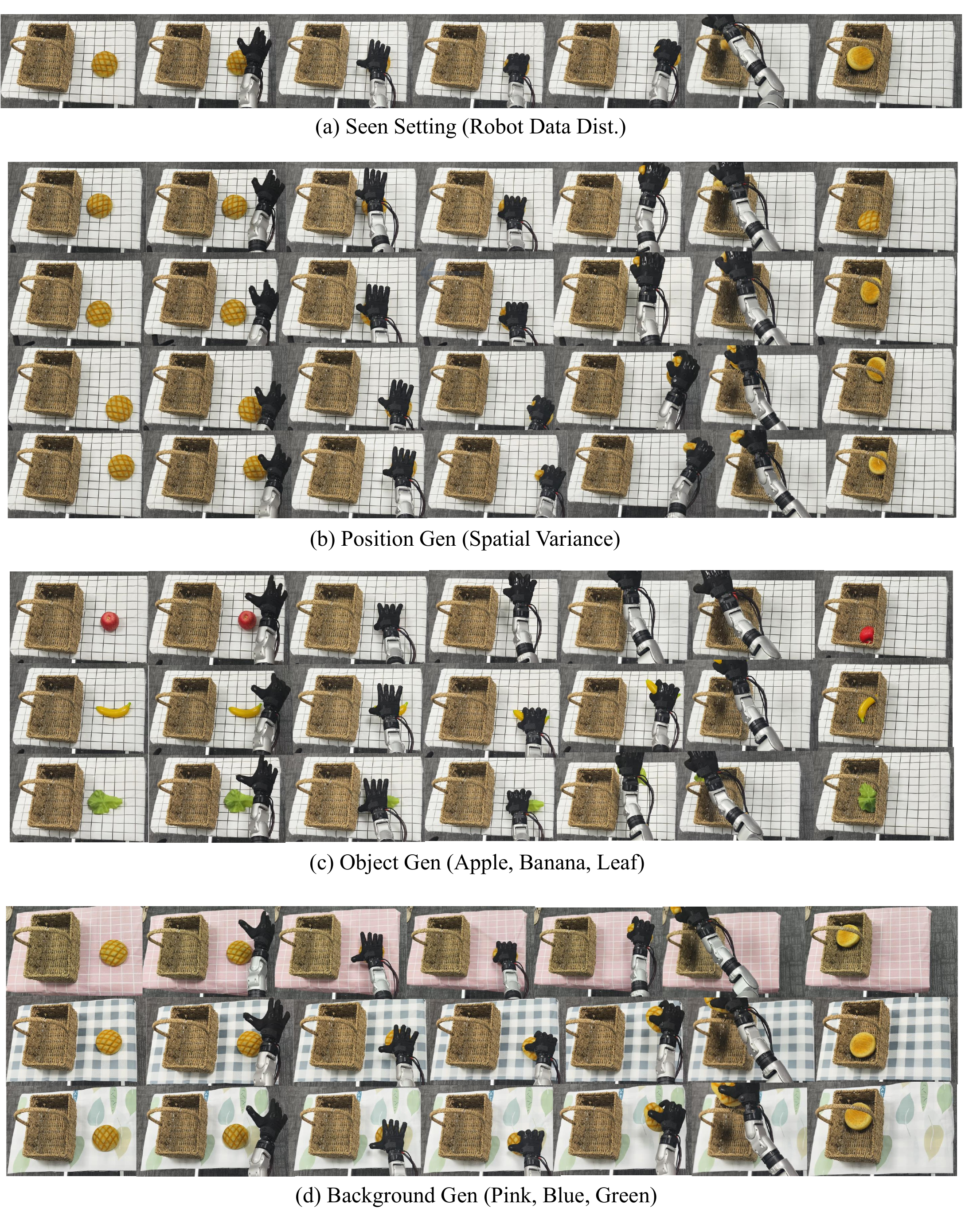}
    \caption{\textbf{Generalization Protocols.} We evaluate the policy under four conditions: 
    (a) \textbf{Seen Setting:} Matches the robot fine-tuning data (fixed setup). 
    (b) \textbf{Position Generalization:} Random initialization within a bounding box, reflecting spatial variance in human data. 
    (c) \textbf{Object Generalization:} Unseen instances (Apple, Banana, Leaf) to test semantic robustness. 
    (d) \textbf{Background Generalization:} Unseen table textures (Pink, Blue, Green) to test visual robustness.}
    \label{fig:gen_settings}
\end{figure*}

~\ref{fig:gen_settings}

\subsection{Generalization Setting Details}
\label{sec:gen_settings}

To evaluate whether the diversity of our large-scale human data is effectively transferred to the robot, we introduce three generalization categories beyond the standard "Seen Setting".
Specifically, the \textbf{Seen Setting} aligns with the constrained distribution of the \textit{Robot Data} (used for fine-tuning), while the \textbf{Generalization Settings} (Position, Object, Background) mimic the extensive variations present in the \textit{Human Data} (used for pre-training). The visual configurations are illustrated in Fig..

\begin{itemize}
    \item \textbf{Seen Setting (Robot Domain):} 
    This setting replicates the exact distribution of the robot fine-tuning data. The object position is fixed (or has negligible variance), and the object instance and background texture remain identical to the specific tasks defined in the fine-tuning stage.

    \item \textbf{Position Generalization:} 
    To verify if the policy inherits spatial awareness from human data, we randomize the target object's position. The object is placed at random coordinates $(x, y)$ within a predefined \textbf{bounding box} (e.g., $22.5cm \times 18.0cm$) on the table surface. This forces the robot to dynamically adapt its approach trajectory rather than overfitting to a fixed motion path.

    \item \textbf{Object Generalization:} 
    We replace the manipulation target with semantically similar but geometrically distinct objects (e.g., replacing a standard bread with an \textbf{Apple}, \textbf{Banana}, or \textbf{Leaf}). This tests whether the representation learned from diverse human interactions allows the robot to handle novel shapes and grasp affordances zero-shot.

    \item \textbf{Background Generalization:} 
    We drastically alter the workspace appearance by covering the table with \textbf{Pink}, \textbf{Blue}, and \textbf{Green} tablecloths. This evaluates the vision encoder's ability—gained from the diverse visual backgrounds in human data—to ignore task-irrelevant shifts and focus on the target object.
\end{itemize}

\end{document}